\documentclass{article}

\usepackage{microtype}
\usepackage{graphicx}
\usepackage{subfigure}
\usepackage{booktabs} 
\usepackage{adjustbox}
\usepackage[table]{xcolor}
\usepackage{multirow}
\usepackage{makecell}
\usepackage{pifont}
\usepackage{enumitem}
\usepackage{chapterbib}
\definecolor{backred}{RGB}{255, 190, 190}
\definecolor{backblue}{RGB}{210, 230, 250}
\newcolumntype{C}[1]{>{\centering\arraybackslash}m{#1}}
\usepackage{hyperref}

\usepackage[accepted]{icml2024}

\usepackage{amsmath}
\usepackage{amssymb}
\usepackage{mathtools}
\usepackage{amsthm}

\usepackage[capitalize,noabbrev]{cleveref}

\theoremstyle{plain}

\theoremstyle{definition}

\theoremstyle{remark}

\usepackage[textsize=tiny]{todonotes}

\icmltitlerunning{SPHINX-X: Scaling Data and Parameters for a Family of Multi-modal Large Language Models}

\begin{document}

\twocolumn[
\icmltitle{SPHINX-X: Scaling Data and Parameters for a Family of\\Multi-modal Large Language Models}



\icmlsetsymbol{equal}{*}

\begin{icmlauthorlist}

\icmlauthor{Dongyang Liu}{equal,cuhk,shlab}
\icmlauthor{Renrui Zhang}{equal,cuhk,shlab}
\icmlauthor{Longtian Qiu}{equal,shlab}
\icmlauthor{Siyuan Huang}{equal,shlab}
\icmlauthor{Weifeng Lin}{equal,shlab}
\icmlauthor{Shitian Zhao}{shlab}
\icmlauthor{Shijie Geng}{rutgers}
\icmlauthor{Ziyi Lin}{cuhk,shlab}
\icmlauthor{Peng Jin}{shlab}
\icmlauthor{Kaipeng Zhang}{shlab}
\icmlauthor{Wenqi Shao}{shlab}
\icmlauthor{Chao Xu}{shlab}
\icmlauthor{Conghui He}{shlab}\\
\icmlauthor{Junjun He}{shlab}
\icmlauthor{Hao Shao}{cuhk}
\icmlauthor{Pan Lu}{ucla}
\icmlauthor{Yu Qiao $^\dagger$}{shlab}
\icmlauthor{Hongsheng Li $^\dagger$}{cuhk,cpii}
\icmlauthor{Peng Gao $^{\dagger\ \ddagger}$}{equal,shlab}

\end{icmlauthorlist}
\icmlaffiliation{shlab}{Shanghai AI Laboratory}
\icmlaffiliation{cuhk}{MMLab, CUHK}
\icmlaffiliation{ucla}{University of California, Los Angeles}
\icmlaffiliation{rutgers}{Rutgers University}
\icmlaffiliation{cpii}{Centre for Perceptual and Interactive Intelligence (CPII)}

\icmlcorrespondingauthor{Peng Gao}{gaopeng@pjlab.org.cn}
\icmlcorrespondingauthor{Hongsheng Li}{hsli@ee.cuhk.edu.hk}
\icmlcorrespondingauthor{Yu Qiao}{qiaoyu@pjlab.org.cn}

\icmlkeywords{Machine Learning, ICML}

\vskip 0.3in
]



\printAffiliationsAndNotice{$^*$ Equal Contribution $^\dagger$ Corresponding Authors $^\ddagger$ Project Lead} 

\begin{abstract}
We propose SPHINX-X, an extensive Multi-modality Large Language Model (MLLM) series developed upon SPHINX. To improve the architecture and training efficiency, we modify the SPHINX framework by removing redundant visual encoders, bypassing fully-padded sub-images with skip tokens, and simplifying multi-stage training into a one-stage all-in-one paradigm. To fully unleash the potential of MLLMs, we assemble a comprehensive multi-domain and multi-modal dataset covering publicly available resources in language, vision, and vision-language tasks. We further enrich this collection with our curated OCR intensive and Set-of-Mark datasets, extending the diversity and generality. By training over different base LLMs including TinyLlama-1.1B, InternLM2-7B, LLaMA2-13B, and Mixtral-8$\times$7B, we obtain a spectrum of MLLMs that vary in parameter size and multilingual capabilities. Comprehensive benchmarking reveals a strong correlation between the multi-modal performance with the data and parameter scales. Code and models are released at \url{https://github.com/Alpha-VLLM/LLaMA2-Accessory}.
\end{abstract}

\section{Introduction}
Since the release of OpenAI's GPT-4~(V)~\cite{openai2023gpt4v} and Google's Gemini~\cite{team2023gemini}, Multi-modal Large Language Models (MLLMs) have become an increasingly popular research area~\cite{fu2023challenger,yang2023dawn}. By aligning multi-modal encoders with Large Language Models (LLMs), MLLMs demonstrate the potential to unlock myriad novel applications and further push the boundary of next-level artificial general intelligence, spanning from embodied intelligence~\cite{geng2023sage}, autonomous driving~\cite{wen2023dilu, cao2023towards,yang2023lidar} to graphical user interfaces (GUI) agents~\cite{he2024webvoyager,yang2023appagent}. 

\begin{figure}[t]
    \centering
    \includegraphics[width=0.95\linewidth]{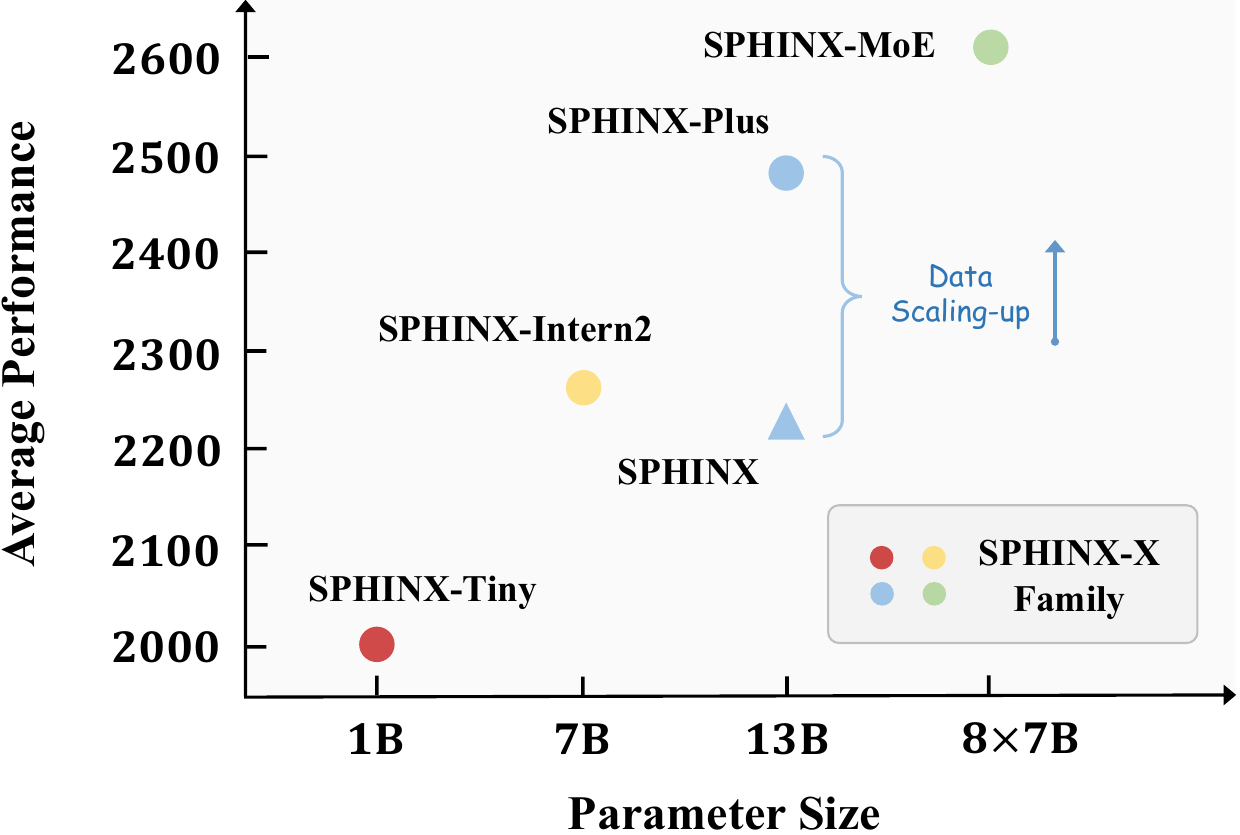}
    \vspace{-6pt}
    \caption{\textbf{Performance comparison with data and parameter scaling.} We introduce SPHINX-X, a general and well-performing MLLM family developed upon SPHINX~\cite{lin2023sphinx}.}
    \label{fig:teaser}
    \vspace{-0.3cm}
\end{figure}

\begin{figure*}[t]
    \centering
    \includegraphics[width=\linewidth]{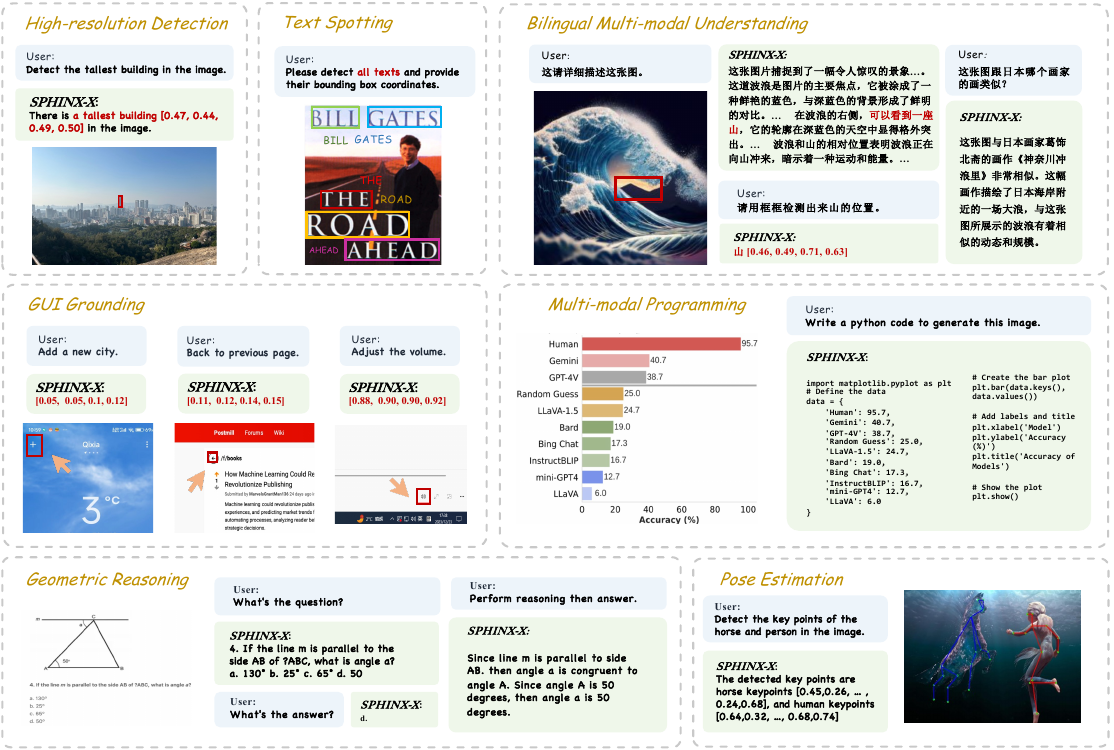}
    \vspace{-0.5cm}
    \caption{\textbf{Demonstrations of SPHINX-X.} With our proposed data and training strategies, SPHINX-X can achieve superior multi-modal understanding and reasoning capabilities in diverse domains, e.g., bilingual, serving as a multi-purpose vision generalist.}
    \label{fig:fig2}
\end{figure*}

Inspired by this, a wide array of open-source MLLMs have been developed within merely one year, including BLIP series~\cite{pmlr-v202-li23q,instructblip}, LLaMA-Adapter~\cite{zhang2024llamaadapter, gao2023llamaadapterv2}, LLaVA~\cite{liu2023llava, liu2023improvedllava,li2024llavanext-strong,li2024llavanext-interleave}, MiniGPT-4~\cite{zhu2023minigpt}, mPLUG-Owl~\cite{ye2023mplugowl, ye2023mplugowl2}, SPHINX~\cite{lin2023sphinx}, and the visual mathematical MAVIS~\cite{zhang2024mavis}.
Although these open-source MLLMs demonstrate promising multi-modal capabilities, their performance is still constrained by the training data from few task domains and limited choices of LLM parameters:

\noindent \underline{\textbf{\textit{Limited Data Coverage for Tasks.}}} Popular open-source MLLMs, such as BLIP-2, LLaVA, and LLaMA-Adapter, are typically trained on raw vision-language data from the natural image domain (e.g., LAION~\cite{schuhmann2021laion, schuhmann2022laion},
SBU~\cite{NIPS2011_5dd9db5e}, and Conceptual Captions~\cite{sharma2018conceptual}),
and visual instruction-following data~\cite{Chen2023ShareGPT4VIL,lavisinstruct} generated by GPT-4~(V)~\cite{openai2023gpt4v}.
As a result, they normally exhibit a superior multi-modal understanding performance in natural images. However, they display limited or degraded results in out-of-domain scenarios, such as Optical Character Recognition (OCR), table, chart, and mathematics fields, where in-depth domain-specific knowledge is critical. In contrast, domain-specific MLLMs like Shikra~\cite{chen2023shikra}, mPLUG-DocOwl~\cite{ye2023mplug}, and Kosmos-2.5~\cite{lv2023kosmos} are tailored to excel in specific tasks, but at the expense of their general multi-modal capabilities.

\noindent \underline{\textbf{\textit{Limited Choices of Model Parameters.}}} Most open-source MLLMs are developed on top of dense LLMs, e.g., LLaMA~\cite{touvron2023llama,touvron2023llama2}, with 7B or 13B parameters. While such parameter counts are often prohibitively large for deployment on portable devices, the same number of parameters remains inadequate to fully explore the performance boundaries of MLLMs. Therefore, scaling down the model scale of MLLMs could facilitate the broader adoption of mobile devices. Meanwhile, scaling up the parameter count through the integration of sparsely-activated Mixture-of-Experts (MoE) architecture~\citep{shazeer2017outrageously} could also unlock the full potential of MLLMs in addressing complex real-world multi-modal challenges.
 

To resolve the aforementioned limitations of existing MLLMs, we introduce a family of MLLMs termed \textbf{SPHINX-X} by extending the data coverage of tasks and parameter scales in SPHINX, as shown in Figure~\ref{fig:teaser}. The superior multi-modal generalization capacity of SPHINX-X for a diversity of tasks is exhibited in Figure~\ref{fig:fig2}. Importantly, we adjust the training process and model architecture of SPHINX to better accommodate the efficient and large-scale multi-modal training:

\noindent \underline{\textbf{\textit{\ding{192} Modifications over SPHINX.}}} ~For the mixed four vision encoders in SPHINX, we only preserve two of them, i.e., CLIP-ConvNeXt~\cite{liu2022convnet} and DINOv2~\cite{oquab2023dinov2}. Considering their distinct methodologies and architectures, the two encoders can provide the most complementary visual semantics, denoted as \textbf{M}ixture \textbf{o}f \textbf{V}isual experts (\textbf{MoV}).
Then, for the sub-image division strategy of high-resolution images, if the input image has a large aspect ratio, we observe a frequent occurrence of fully-padded sub-images, where all pixels are zeros. To address this, we adopt a learnable skip token to represent them within LLMs, thereby shortening the sequence length for efficiency, while still preserving the relative positions of sub-images.
Furthermore, given the increased training data volume, we condense the previous multi-stage training pipeline into a more straightforward single-stage paradigm.
Instead of fine-tuning different parts of LLM parameters in two stages with different datasets, we directly train all the parameters of LLMs on all our collected datasets.

\noindent \underline{\textbf{\textit{\ding{193} Multi-Domain and Multi-Modal Datasets.}}} ~To fully unleash the potential of MLLMs, we assemble an extensive collection of public datasets that span a wide array of tasks, and carefully extend two self-curated multi-modal datasets. In detail, we collect the public datasets from the realms of vision, language, and vision-language tasks, and reformulate them into a unified multi-turn conversational format. 
%
Moreover, to specifically enhance the targeted capacity of MLLMs, we further construct an OCR-intensive dataset and a Set-of-Mark (SoM) dataset. 
The expansion of OCR data processed from substantial PDFs can unlock the visual language understanding power of MLLMs, e.g., text spotting and document layout detection.
The specialized SoM data also compensates for the SoM prompting \cite{yang2023set} potentials of SPHINX-X, for which we construct delicate SoM annotations in diverse domains by GPT-4.


\noindent \underline{\textbf{\textit{\ding{194} LLM Parameter Scaling of SPHINX-X.}}} 
~With the aforementioned techniques and large-scale
datasets, we marry SPHINX-X with various base LLMs of increasing parameter scales: TinyLlama-1.1B~\cite{zhang2024tinyllama}, InternLM2-7B~\cite{team2023internlm},
LLaMA2-13B~\cite{touvron2023llama2}, and Mixtral-8×7B~\citep{jiang2024mixtral}.
Respectively, we develop a family of MLLMs that facilitate fast mobile deployment (SPHINX-Tiny), provide bilingual support (SPHINX-Intern2), possess moderate parameters with data scaling (SPHINX-Plus), and exhibit strong reasoning capabilities through Mixture-of-Expert architectures (SPHINX-MoE). 

Extensive evaluations across a wide range of benchmarks reveal that SPHINX-Plus surpasses the original SPHINX, confirming that enriching dataset scales and diversity can benefit the performance. 
Furthermore, a comparison of base LLMs from 1.1B to 7$\times$8B demonstrates that under the same training pipeline, scaling up the parameters can consistently boost the multi-modal understanding capabilities. Overall, we summarize the key contributions as follows:
\vspace{-10pt}
\begin{itemize}[leftmargin=*,noitemsep]
    \item We release a family of well-performing MLLMs tailored from fast inference on mobile devices to complex reasoning tasks on high-end computers. A comprehensive range of experiments demonstrates that the scale of training data and the size of LLM parameters both play a critical role in the performance of MLLMs. 

    \item We perform several modifications over SPHINX by eliminating redundant visual encoders, avoiding fully-padded sub-images with learnable skip tokens, as well as streamlining the complex multi-stage training pipeline into a single-stage all-in-one paradigm. 

     \item We collected an extensive multi-modal dataset covering a broad spectrum of tasks and modalities. On top of that, we curated two new datasets for enhancing the OCR-intensive and Set-of-Marks prompting capabilities of MLLMs.
\end{itemize}

\section{Related Work}


\paragraph{Large Language Models (LLMs)}
Advancements in recent MLLM research are based on the breakthrough of LLMs constructed upon the Transformer architecture~\citep{vaswani2017attention}, where progress has stemmed from both an expansion of training data and a significant increase in model parameters. For instance, GPT-3~\citep{brown2020language}, boasting 175B parameters, excels at few-shot in-context learning, while GPT-2~\citep{radford2019language} with 1.5B parameters falls short of reaching this level of performance. 
Inspired by GPT-3's success, several LLMs like PaLM~\citep{chowdhery2022palm}, OPT~\citep{zhang2022opt}, BLOOM~\citep{workshop2022bloom}, and LLaMA have emerged.
Mistral~\citep{jiang2023mistral} further introduced window attention for enhanced long-context modeling, while Mixtral 8$\times$7B leveraged sparse MoE layers~\citep{fedus2022switch,lepikhin2020gshard,shazeer2017outrageously} to upscale parameters efficiently, outperforming with fewer active parameters. Concurrently, models such as Qwen~\citep{bai2023qwen}, Baichuan~\citep{yang2023baichuan}, and InternLM~\citep{team2023internlm} have advanced bilingual LLM capabilities, whereas TinyLlama~\citep{zhang2024tinyllama} and Phi-2~\citep{phi2} focus on reducing parameters for edge deployment. Our SPHINX family extends LLMs to multimodal domains for visual understanding and reasoning. We select four LLMs with different pre-training and parameter scales, comparing their performance under multi-modal scenarios.
\paragraph{Multi-modal Large Language Models (MLLMs)}
Efforts to extend LLMs to perceive beyond text have birthed MLLMs, with vision as the primary modality. Representative architectures include BLIP series~\citep{li2022blip,pmlr-v202-li23q} and MiniGPT-4~\citep{zhu2023minigpt}, which employ query Transformers to summarize visual features; Flamingo~\citep{alayrac2022flamingo}, which uses gated cross-attention to support interleaved image-text inputs; The LLaMA-Adapter series~\citep{zhang2024llamaadapter,gao2023llamaadapterv2} that introduce zero-initialized attention to minimize interference between visual and language tokens; and LLaVA~\cite{liu2023llava,liu2023improvedllava}, which connects visual tokens to LLMs with a simple linear layer to incorporate visual knowledge. 
There are also recent advances in MLLMs that have demonstrated remarkable extended capabilities. For example, Shikra~\citep{chen2023shikra} and VisionLLM~\citep{wang2023visionllm} excel in referring object detection, while ChartAssistant~\citep{meng2024chartassisstant} and mPLUG-DocOwl/PaperOwl~\citep{ye2023mplug,hu2023mplug} focus on tables, documents, and scientific diagrams analysis. LLaVA-NeXT-Interleave~\cite{li2024llavanext-interleave} targets multi-image interleaved instruction tuning, and MAVIS~\cite{zhang2024mavis} curates large-scale visual mathematical tuning datasets for challenging problem solving. Many efforts also extend LLMs into more modalities, such as ImageBind-LLM ~\citep{han2023imagebind}, Point-LLM~\cite{guo2023point}, and others~\cite{zhu2023pointclip,zhang2022pointclip,zhang2023prompt}. In this paper, we upgrade SPHINX~\cite{lin2023sphinx} to an MLLM family for more general visual instruction following, achieving superior performance over various benchmarks.



\begin{figure*}[t]
    \centering
    \vspace{0.2cm}
    \includegraphics[width=0.8\linewidth]{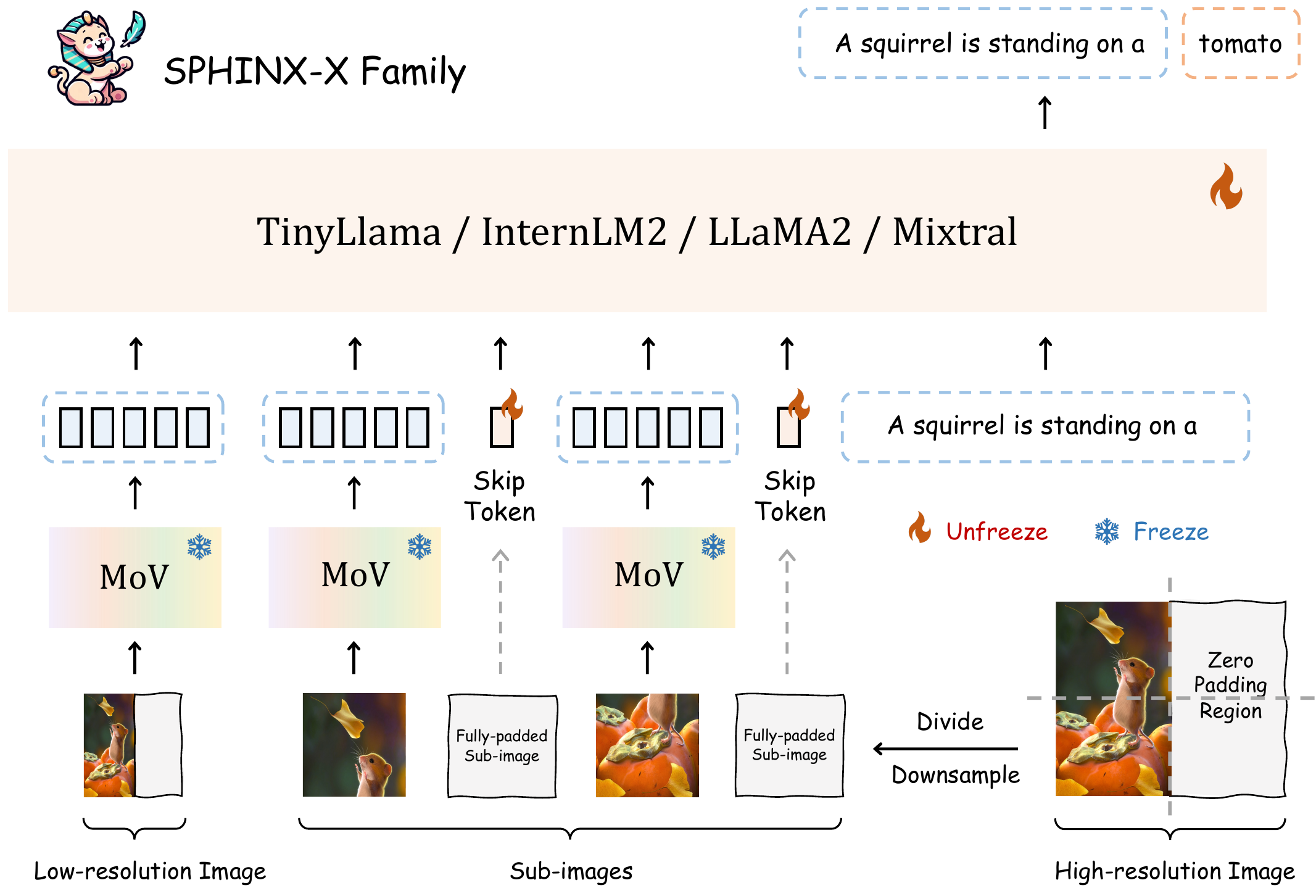}
    \caption{\textbf{Overall paradigm of SPHINX-X family.} On top of SPHINX~\cite{lin2023sphinx}, we adopt three modifications for a more general and concise architecture: removing redundant visual encoders in Mixture of Visual Experts (MoV), bypassing fully-padded sub-images with skip tokens, and simplifying multi-stage training into a one-stage all-in-one approach.}
    \label{fig:arch}
\end{figure*}

\section{Method}
We first revisit the design principles of SPHINX in Section~\ref{sec:revisit}. We then respectively detail the three improvements made to SPHINX-X in Section~\ref{sec:improve} concerning the succinctness of visual encoders, learnable skip tokens for useless visual signals, and simplified one-stage training. 
Lastly, we illustrate the composition of our large-scale multi-modality dataset in Section~\ref{sec:train data}, as well as introduce different base LLMs adopted by the SPHINX-X family in Section~\ref{sec:llm}.


\subsection{A Revisit of SPHINX}
\label{sec:revisit}
SPHINX~\cite{lin2023sphinx} proposes three types of mixing strategies to develop a multi-purpose MLLM -- mixing of model weights, tuning tasks, and visual embeddings. 
Following previous efforts~\cite{gao2023llamaadapterv2,liu2023llava}, SPHINX adopts a two-stage training pipeline, in which the first stage aligns pre-trained vision encoders with LLaMA2~\cite{touvron2023llama2}, and the second stage integrates a variety of tasks for instruction tuning. 
For more robust visual representations, SPHINX incorporates the embeddings of four different vision encoders, including CLIP-ViT~\cite{Radford2021LearningTV,Dosovitskiy2020AnII}, CLIP-ConvNeXt~\cite{liu2022convnet}, DINOv2~\cite{oquab2023dinov2}, and Q-former~\cite{Li2023BLIP2BL}. SPHINX then introduces a multi-scale mixing strategy to tackle high-resolution images, which divides the high-resolution input into several sub-images along with a downsampled image for concurrent encoding. 
In addition, to further mix various domain semantics, SPHINX fuses the first-stage weights of LLMs that are tuned by different data domains. 
Despite its superior performance, SPHINX is still constrained by the cumbersome two-stage training process and mixed architectures, and it has yet to fully capitalize on the potential benefits of data and model scaling. 
Motivated by this, we develop SPHINX-X, an extensive series of MLLMs to explore a more general and comprehensive multi-modal learning paradigm. 


\subsection{SPHINX-X}
\label{sec:improve}
To better handle large-scale multi-task and multi-modal instruction-tuning, we perform the following improvements over SPHINX-X, enabling the training pipeline and model architecture to be concise.~We present the upgraded SPHINX-X training pipeline in Figure \ref{fig:arch}.

\paragraph{Eliminating Redundant Visual Encoders.}
SPHINX employs four complementary vision encoders to capture diverse visual representations. Although the mixture of visual experts can improve the performance, it inevitably leads to a significant increase in computational costs, especially for a group of sub-images generated from a high-resolution input. To obtain better computational efficiency, we eliminate the CLIP-ViT and Q-former encoders, only preserving two visual encoders -- DINOv2 and CLIP-ConvNeXt. 
As these two models are pre-trained by distinct learning approaches (self-supervised vs. weakly-supervised) and network architectures (ViT vs. CNN), they can already provide the most complementary and refined visual knowledge. We denote them as the \textbf{M}ixture \textbf{o}f \textbf{V}isual experts (\textbf{MoV}).

\paragraph{Bypassing Fully-padded Sub-images with Skip Tokens.}
The superior performance of SPHINX can be attributed to  its effective handling of high-resolution images with several local sub-images and one global downsampled image. During the training stages of SPHINX, all images by default are scaled and zero-padded to a high resolution $448 \times 448$, and then divided into four $224 \times 224$ sub-images. 
However, for images with large aspect ratios, say $2 : 1$, this operation will result in fully-padded sub-images filled entirely with zero-value pixels. 
Such fully-padded sub-images not only contain noisy and useless visual signals, but also produce spare visual tokens that waste computational resources within both MoV and LLM.
To alleviate the issue, we propose a learnable skip token to replace the fully-padded sub-image, which provides explicit relative positional information for LLMs to identify the positions between useful sub-images. 
In this way, MoV can avoid encoding these zero-pixel sub-images, which allows for a reduction in the input sequence length for LLMs, achieving enhanced computational efficiency.

\paragraph{One-Stage All-in-One Training.}
The original training pipeline of SPHINX comprises two stages and utilizes a weight mixing strategy. 
However, it requires to manually assign various tunable parameters and dataset combinations to different training stages, which is a labor-intensive task.
To simplify the overall paradigm, we design a single-stage all-in-one training pipeline, which treats all collected datasets equally and uniformly transforms them into multi-modal multi-turn dialog formats. 
During the one-stage training, we unfreeze all the parameters of SPHINX (i.e., LLM and intermediate projection layers) except for the two visual encoders in MoV. Due to the large volume of training data and high reasoning capacity of LLMs, our one-stage all-in-one strategy can significantly streamline the training procedure for MLLMs while maintaining high performance.

\subsection{Training Data of SPHINX-X}
\label{sec:train data}

To obtain remarkable multi-modal capabilities, we widely convert three categories of public training data into instruction-following formats (language, vision, and vision-language), and carefully curate two targeted multi-modal datasets (OCR-intensive and Set-of-Mark) for SPHINX. All data is combined for the one-stage all-in-one training. Specifically, for natural language data, we utilized datasets that include multi-turn dialog, mathematical reasoning, and code generation. For vision data, we convert data from diverse computer vision tasks including image-level and object-level understanding into multi-turn conversation formats. For vision-language data, we collect various visual question-answering, visual instruct-tuning, and fine-grained image captioning datasets. On top of this, we generate an OCR dataset from large-scale PDF data, and a multi-domain Set-of-Marks dataset with fine-grained multi-modal knowledge.

For the three parts that are mainly composed of existing datasets (namely {\it language instruction-following, visual instruction-following and vision-language instruction-following}), we defer the details to the appendix (Sec. \ref{sec:train_data_appendix}). The data statistic information is also provided in the appendix (Table~\ref{tab:data_summary}). 

\noindent \textbf{OCR-intensive Data.}
Most previous MLLMs can only leverage external tools and pre-extracted OCR tokens to obtain satisfactory OCR-related understanding.
To enhance such capabilities for MLLMs, we compile an OCR-intensive dataset from extensive Internet PDF data. Different from previous synthetic OCR data~\cite{yim2021synthtiger, kim2021donut} that are too simple and far from real-world application, our dataset is more challenging and larger-scale.
Specifically, we first collect large-scale PDF datasets from Common Crawl~\footnote{\href{https://commoncrawl.org/}{Common Crawl:https://commoncrawl.org/}} and arXiv websites. Then, we utilize PyMuPDF~\footnote{\href{https://github.com/pymupdf/PyMuPDF}{PyMuPDF: https://github.com/pymupdf/PyMuPDF}} to get the rendering results of each page in the PDF file and also save all the text annotations along with their bounding boxes. To ensure the OCR quality, we adopt multiple processing methods, including Unicode characteristic checking, text splits merge, etc. In this way, we constructed an in-house PaperText dataset with about 3M text-dense pages. Finally, we transform them into a unified question-answering format to strengthen the OCR documentation understanding ability. 



\noindent \textbf{Multi-Domain Set-of-Mark Data.}
We notice that existing multi-modal datasets lack the fine-grained correspondence between images and texts. Thus, we construct a multi-domain dataset similar to Set-of-Marks techniques~\cite{yang2023set} to endow MLLMs with dense multi-modal captioning knowledge.
Initially, we collect diverse image datasets from various domains. Then, we utilize dataset annotations such as bounding boxes and object masks to place various marks like points, boxes, polygons, and identifiers, on the raw images.
After that, we craft domain-specific instructions for each data type, and prompt GPT-4V with the masked images for multi-scale captioning, which generates captions of global image understanding, detailed region captioning, and object-relation analysis. Such SoM prompting for GPT-4V can motivate its power to produce higher-quality and fine-grained multi-modal data. 
During training, we do not utilize the marked images, but the raw images, and describe the marks by language within the multi-turn conversations for uniformity with other data domains.

\subsection{SPHINX-X with Different LLMs}
\label{sec:llm}

Built upon the aforementioned techniques and large-scale datasets, we provide four choices of base LLMs in SPHINX-X with increasing parameter scales: TinyLlama-1.1B~\cite{Zhang2024TinyLlamaAO}, InternLM2-7B~\cite{team2023internlm}, LLaMA2-13B~\cite{touvron2023llama2}, and Mixtral-8$\times$7B~\cite{jiang2024mixtral}. We introduce their features compared to the original SPHINX with LLaMA2-13B.

\noindent \textbf{SPHINX-Tiny} with TinyLlama-1.1B. TinyLlama can be regarded as a lightweight version of LLaMA.
The compactness of 1.1B parameters allows TinyLlama to apply to a diversity of scenarios with limited computation resources. Therefore, we train SPHINX-Tiny to observe how the multi-modal performance varies given the smaller-scale LLM.

\begin{table*}[t]
\centering
\caption{\textbf{Performance comparison with state-of-the-art methods on popular MLLM benchmarks.}}
\vspace{2mm}
\label{table:mm}
\adjustbox{max width=\textwidth}{%
\setlength{\tabcolsep}{3pt}
\begin{tabular}{@{}l|ccccccccccccc@{}}
\toprule
Methods & POPE & MME\textsuperscript{P} & MME\textsuperscript{C} & MMB & SEED & LLaVA\textsuperscript{W} & MM-Vet & CCbench & MathVista  & Tiny LVLM & BenchLMM & InfiMM-Eval & Qbench \\ 
\midrule
BLIP-2                    & 85.3          & 1293.8          & -               & -             & 46.4           & 38.1          & 22.4          & -             & -      & 284.7 &  -  &  - &  -    \\
InstructBLIP-7B             & -             & -               & -               & 36.0 & 53.4           & 60.9          & 26.2          & 12.1          & 25.3  & 300.6 & 44.63  &  - & 56.7     \\
InstructBLIP-13B            & 78.9          & 1212.8          & -             & -             & -              & 58.2          & 25.6          & -             & -    &  - &  45.03   &  - &  -       \\
LLaMA-AdapterV2           & -             & 1328.4         & 356.4          & - & -              & -             & -             & -             & -  & 229.2 & - & 30.5   & 59.5      \\
Qwen-VL-7B                  & -             & -               & -               & 38.2          & 56.3           & -             & -             & 5.5           & -    & -  & - &  - & 59.4         \\
Qwen-VL-7B-Chat            & -             & 1487.6         & 360.7 & 60.6 & 58.2           & -             & -             & \textbf{39.3} & -    & 316.8 & - & 37.4 & -        \\
LLaVA1.5-7B                 & 85.9          & 1510.7          & -               & 64.3 & 58.6           & 63.4          & 30.5          & 16.4          & -   & - & 46.8  & - & 58.7  \\
LLaVA1.5-13B                 & 85.9          & 1531.3          & 295.4          & 67.7 & 61.6           & 70.7          & 35.4          & 26.5          & -    & 307.2 & 55.5 & 32.62  &  62.1       \\
SPHINX                 & \textbf{90.8} & \textbf{1560.2} & 310.0           & 67.1   & 71.6  & 74.3           & 36.6 & 27.9          & 27.5 & 288.9 & - & 30.7 & 65.8 \\
\midrule
\rowcolor[gray]{0.95}
SPHINX-Tiny                 &    82.2        &   1261.2         &   242.1        &       56.6         & 17.1 & 52.3  &  23.8              &  17.5       & 26.4 & 301.5 & 50.0 & 21.9 & 19.7 \\
\rowcolor[gray]{0.95}
SPHINX-Intern2                 & 86.9           &  1260.4           &  294.6      &   57.9             & 68.8  & 57.6  & 36.5               &   21.0    & 35.5  &312.9 & 47.0 & 31.5 & 60.0 \\
\rowcolor[gray]{0.95}
SPHINX-Plus                 &   89.1        &  1457.7          &    283.6      &     71.0           & \textbf{74.8} & 71.7  &    \textbf{47.9}            &  25.6       & 36.8  & 282.1 & \textbf{57.4} & \textbf{39.5} & \textbf{68.6} \\
\rowcolor[gray]{0.95}
SPHINX-MoE                &      89.6      &     1485.3       &    \textbf{367.1}       &  \textbf{71.3}            & 73.0 & 70.2  &  40.9             &  15.4        & \textbf{42.7}  & \textbf{335.3} & 50.7 & 38.6 & 66.2 \\
\bottomrule
\end{tabular}
} 
\vspace{-3mm}
\end{table*}

\noindent \textbf{SPHINX-MoE} with Mixtral-8$\times$7B. 
As a sparse Mixture-of-Experts (MoE) LLM, Mixtral-8$\times$7B utilizes 8 feed-forward networks at each transformer layer as experts, and relies on a router network to activate two experts each time. 
With this sparse mechanism, we expect to analyze the characteristics of different experts for multi-modal instruction following.

\noindent \textbf{SPHINX-Plus} with LLaMA2-13B.
SPHINX-Plus utilizes the same scaled LLaMA, i.e., 13B parameters, with the original SPHINX, but is tuned by our constructed multi-modal dataset (more diverse and larger-scale) with one stage. 
This to some extent can illustrate the efficacy of data scaling-up. Note that, referring to SPHINX-2K~\cite{lin2023sphinx}, we also perform an improved version, termed \textbf{SPHINX-Plus-2K}, which increases the image resolution from $448\times448$ to $672\times672$ and splits the input to $3\times3$ sub-images for fine-grained visual understanding. 

\noindent \textbf{SPHINX-Intern2} with InternLM2-7B.
InternLM~\cite{team2023internlm} is a strong bilingual LLM pre-trained on large Chinese and English corpus. The recently released InternLM2-7B shows stronger bilingual language understanding ability.
We adopt InternLM2-7B as the base LLM to explore the performance pattern under the regular 7B-parameter LLM setup and the potential of bilingual multi-modal reasoning.

\section{Experiment}
\label{sec:experiments}

\subsection{Experimental Settings}
All SPHINX-X models presented in the paper follow the one-stage all-in-one training strategy, and all modules except the visual encoders are optimized. The learning rate is set to 5e-6 for SPHINX-MoE, and 2e-5 for others. During training, the learning rate first linearly warmups to the target value within the first 0.01 epoch, and then gradually decays to 0 following the cosine schedule. We use the AdamW optimizer with weight decay = $0$ and betas = $(0.9, 0.95)$. To accommodate the large model volume, a combination of ZeRO2-style~\cite{zero} data parallel and Megatron-style~\cite{megatron} model parallel is used. The model parallel size is set to 8 for SPHINX-MoE, 2 for SPHINX-Plus, and 1 for others. The effective batch size is 256. Note that SPHINX-Plus is initialized from the SPHINX model, while other SPHINX-X models use the original visual encoders/LLMs and randomly initialized linear projection layers.
In Figure~\ref{fig:teaser}, we report the cumulative scores of SPHINX-X models over benchmarks mentioned in Tables~\ref{table:mm}, \ref{table:textvqa}, and \ref{table:ref}. However, due to the excessively high overall score, we excluded MME~\cite{fu2023mme} perception to balance the results.

\begin{table}[t]
\centering
\vspace{-2mm}
\caption{\textbf{Performance on 7 academic VQA benchmarks.}}
\vspace{2mm}
\label{table:vqa}
\scriptsize
\setlength{\tabcolsep}{2pt}
\begin{tabular}{l|cccccc}
\toprule
\multicolumn{1}{l|}{\multirow{1}{*}{Method}} & OKVQA          & VQAV2         & VizWiz        & GQA                    & SQA     & IVQA                                                 \\ 
\midrule
BLIP-2                                     & 45.9           & -             & 19.6          & 41.0                  & -             & 40.6                                                         \\
InstructBLIP                                  & -              & -             & 33.4          & 49.5                 & -             & 44.8                                                          \\
LLaMA-AdapterV2                               & 49.6          & 70.7           & 39.8          & 45.1                    & -             & -                                                                \\
Shikra                                        & 47.2           & 77.4          & -             & -                      & -             & -                                                              \\
Fuyu-8B                                      & 60.6           & 74.2            & -           & -                 & -             & -                                                       \\
MiniGPT-v2                                    & 57.8           & -             & 53.6 & 60.1                   & -             & 51.5                                                            \\
Qwen-VL-7B                                    & 58.6           & 79.5          & 35.2          & 59.3                   & 67.1          & -                                                 \\
Qwen-VL-7B-Chat                               & 56.6           & 78.2          & 38.9          & 57.5                   & 68.2          & -                                                             \\
LLaVA1.5-7B                                   & -              & 78.5          & 50.0            & 62.0                     & 66.8          & -                                                              \\
LLaVA1.5-13B                                   & -              & 80.0            & 53.6 & 63.3        & 71.6 & -                                                  \\ 
SPHINX                                  & 62.2 & 80.2 & 46.8         & 62.9        & 69.1         & 52.7   \\
\rowcolor[gray]{0.95}
\midrule
SPHINX-Tiny                                     &53.6	&	74.7 & 49.2&	58.0	&21.5         & 40.7   \\ 
\rowcolor[gray]{0.95}
SPHINX-Intern2                                  &55.5	&  75.5	&49.6	&56.2	&70.4	&49.0   \\ 
\rowcolor[gray]{0.95}
SPHINX-Plus                                   & - &	- & 	57.8 &	-	&	74.2&	54.7 \\
\rowcolor[gray]{0.95}
SPHINX-MoE                                  &\textbf{62.7}	&  \textbf{81.1}	&\textbf{61.9}	&\textbf{63.8} &	\textbf{74.5}&	\textbf{57.3} \\ 
\bottomrule
\end{tabular}
\vspace{-2.5mm}
\end{table}

\subsection{Performance Evaluation}
In this section, we conduct a thorough assessment and present outcomes across various benchmarks, offering an extensive overview and evaluation of our SPHINX-X family.

\begin{table*}[!t]
\small
\centering
\caption{\textbf{Evaluation results of SPHINX-MoE on MathVerse~\cite{zhang2024mathverse} and SciVerse~\cite{sciverse}.}}
\vspace{2mm}
\begin{adjustbox}{width=\linewidth}
    \begin{tabular}{l|ccccccc|ccccccc}
    \toprule
    \multirow{4}*{\makecell*[l]{\large Model}}  
    &\multicolumn{7}{c|}{\large MathVerse} &\multicolumn{7}{c}{\large SciVerse} \\  
    &\multicolumn{1}{c}{\makecell*[c]{All}}
    &\multicolumn{1}{c}{\makecell*[c]{\shortstack{\vspace*{0.1pt}\\Text\\\vspace*{0.2pt}\\Dominant}}} 
    &\multicolumn{1}{c}{\makecell*[c]{\shortstack{\vspace*{0.1pt}\\Text\\\vspace*{0.2pt}\\Lite}}}
    &\multicolumn{1}{c}{\makecell*[c]{\shortstack{\vspace*{0.1pt}\\Text\\\vspace*{0.2pt}\\Only}}}
    &\multicolumn{1}{c}{\makecell*[c]{\shortstack{\vspace*{0.1pt}\\Vision\\\vspace*{0.2pt}\\Intensive}}}
    &\multicolumn{1}{c}{\makecell*[c]{\shortstack{\vspace*{0.1pt}\\Vision\\\vspace*{0.2pt}\\Dominant}}}
    &\multicolumn{1}{c}{\makecell*[c]{\shortstack{\vspace*{0.1pt}\\Vision\\\vspace*{0.2pt}\\Only}}}
    &\multicolumn{1}{|c}{\makecell*[c]{All}}
    &\multicolumn{1}{c}{\makecell*[c]{\shortstack{\vspace*{0.1pt}\\Text\\\vspace*{0.2pt}\\Only}}}
    &\multicolumn{1}{c}{\makecell*[c]{\shortstack{\vspace*{0.1pt}\\Knowledge\\\vspace*{0.2pt}\\Lite}}} 
    &\multicolumn{1}{c}{\makecell*[c]{\shortstack{\vspace*{0.1pt}\\Knowledge\\\vspace*{0.2pt}\\Rich}}} 
    &\multicolumn{1}{c}{\makecell*[c]{\shortstack{\vspace*{0.1pt}\\Knowledge\\\vspace*{0.2pt}\\Professional}}}
    &\multicolumn{1}{c}{\makecell*[c]{\shortstack{\vspace*{0.1pt}\\Vision\\\vspace*{0.2pt}\\Dominant}}}
    &\multicolumn{1}{c}{\makecell*[c]{\shortstack{\vspace*{0.1pt}\\Vision\\\vspace*{0.2pt}\\Only}}}\\
    \midrule
    \multicolumn{15}{c}{\textit{Open Source MLLM}}\\
    \cmidrule{1-15}
    LLaMA-Adapter V2 &5.8&7.8&6.3&3.9&6.2&4.5&4.4 &9.3 &9.4 &10.7 &11.2 &11.9 &11.8 &10.4 \\
    ImageBind-LLM &10.0&13.2&11.6&12.9&9.8&11.8&3.5 &27.2 &23.6 &27.8 &28.1 &28.2 &28.9 &23.2 \\
    mPLUG-Owl2 &10.3&11.6&11.4&13.8&11.1&9.4&8.0 &- &- &- &- &- &- &- \\
     MiniGPT-v2 &10.9&13.2&12.7&15.3&11.1&11.3&6.4&30.0&28.5&31.4&29.6&30.5&31.6&26.9 \\
     LLaVA-NeXT &15.6&19.4&15.2&18.1&16.8&15.2&11.3&34.9&33.4&34.0&36.8&37.2&35.1&31.3  \\
    \cmidrule{1-15}
    \multicolumn{15}{c}{\textit{Closed Source MLLM}}\\
    \cmidrule{1-15}
    Qwen-VL-Plus &11.8&15.7&11.1&14.5&9.0&13.0&10.0 &- &- &- &- &- &- &- \\
    Gemini-Pro &24.1&26.3&23.5&27.3&23.0&22.3&22.2 &- &- &- &- &- &- &- \\
    Qwen-VL-Max &25.9&30.7&26.1&28.9&24.1&24.1&21.4  &- &- &- &- &- &- &- \\
    GPT-4V &\textbf{41.0}&\textbf{54.7}&\textbf{41.4}&\textbf{48.7}&\textbf{34.9}&\textbf{34.4}&\textbf{31.6} &- &- &- &- &- &- &- \\
    \midrule
    \rowcolor[gray]{0.95}SPHINX-MoE &15.6&22.2&16.4&18.3&14.8&12.6&9.1&\textbf{37.3}&\textbf{41.1}&\textbf{38.9}&\textbf{38.8}&\textbf{41.3}&\textbf{36.3}&\textbf{31.4}  \\
    \bottomrule
    \end{tabular}
\end{adjustbox}
\label{tab:ms}
\end{table*}

\begin{table}[t]
\centering
\caption{\textbf{Performance on text-oriented VQA tasks.} `${\dagger}$' denotes to use ground-truth OCR tokens during inference and training.}
\label{table:textvqa}
\vspace{2mm}
\scriptsize
\setlength{\tabcolsep}{1pt}
\begin{tabular}{l|ccccccccccc}
\toprule
Method    & 
\begin{tabular}{@{}c@{}}Text\\VQA\\ \end{tabular}  & 
\begin{tabular}{@{}c@{}}OCR\\VQA\\ \end{tabular} & 
\begin{tabular}{@{}c@{}}Doc\\VQA\\ \end{tabular} &
\begin{tabular}{@{}c@{}}Chart\\QA\\ \end{tabular} &
\begin{tabular}{@{}c@{}}AI\\2D\\ \end{tabular}  &
\begin{tabular}{@{}c@{}}Deep\\Form\\ \end{tabular} & 
\begin{tabular}{@{}c@{}}Info\\VQA\\ \end{tabular}  & KLC &   WTQ &
\begin{tabular}{@{}c@{}}Tab\\Fact\\ \end{tabular}     & 
\begin{tabular}{@{}c@{}}Visual\\MRC\\ \end{tabular}           \\ 
\midrule

\multicolumn{12}{c}{\it Specialist models} \\

\midrule
Donut                                      & 43.5  & - & 67.5 & 41.8 & - & \textbf{61.6} & 11.6 & 30.0 & 18.8 & 54.6 & 93.9 \\
UReader                                    &57.6 & -& 65.4 & 59.3 & - & 49.5 & \textbf{42.2} & \textbf{32.8} & 29.4 & \textbf{67.6} & \textbf{221.7}\\
\midrule

\multicolumn{12}{c}{\it Generalist models} \\

\midrule
BLIP-2                                                 & 42.5$^{\dagger}$             & 40.6  & - & - & -   & -& - & - & - & -  & -                      \\
InstructBLIP                                                      & 50.7$^{\dagger}$             & 44.8 & - & - & - & -& - & - & - & -  & -                                 \\
LLaMA-AdapterV2                                                                 & 37.4          & - & - & - & -   & -& - & - & - & -  & -                                  \\
Qwen-VL-7B                                                                   & 63.8 & \textbf{75.7}   & 65.1 & \textbf{65.7} & 62.3   & -& - & - & - & -  & -                    \\
Qwen-VL-7B-Chat                                                            & 61.5          & 70.5  & 62.6 & 62.6 & 62.6     & -& - & - & - & -  & -                              \\
LLaVA1.5-7B                                                               & 58.2          & -& - & - & -   & -& - & - & - & -  & -                                    \\
LLaVA1.5-13B                                                           & 61.3          & -   & - & - & -   & -& - & - & - & -  & -                                 \\ 
SPHINX                                 & 58.8         & 70.0 & 35.8 & 22.5 & 38.1 & 0	&24.0&	0&	13.8	&52.9	&95.3 \\
\rowcolor[gray]{0.95}
\midrule
SPHINX-Tiny                                       &57.8	&60.3&	53.0	&34.1	&24.6&	11.8	&26.3&	22.2	&15.3&	51.1&	147.5  \\ 
\rowcolor[gray]{0.95}
SPHINX-Intern2                                 &58.1&	53.0	&56.3	&39.7&	\textbf{63.0} 	&6.5&	31.6&	10.5&	21.1 &	51.4 &	149.3    \\ 
\rowcolor[gray]{0.95}
SPHINX-Plus                                &	65.7&	70.1&	61.2&	53.4	&46.0  &9.2	 &34.7 &	23.9 &	27.1 &	52.8 &	171.0 \\ 
\rowcolor[gray]{0.95}
SPHINX-Plus-2K                                & \textbf{70.6}	&68.9	&\textbf{71.6}&	55.1&	47.4	&23.2	&39.1&	31.1	&\textbf{31.1}	&54.0	&178.4 \\ 
\rowcolor[gray]{0.95}
SPHINX-MoE                                 &68.0	&64.8	&68.4	&55.0	&55.6	&20.7	&41.8	&25.5	&29.9 &	52.7	&184.4 \\ 
\bottomrule
\end{tabular}
\end{table}

\begin{table}[t]
\caption{{\textbf{Performance (Top-1 Accuracy@0.5) on Referring Expression Comprehension (REC) tasks.}}}
\vspace{2mm}
\adjustbox{max width=\linewidth}{%

\setlength{\tabcolsep}{3.5pt}
\begin{tabular}{@{}l|ccc|ccc|cc@{}}
\toprule

\multirow{2}{*}{Method} & \multicolumn{3}{c|}{RefCOCO+} & \multicolumn{3}{c|}{RefCOCO} & \multicolumn{2}{c}{RefCOCOg} \\
 & val & test-A & test-B & val & test-A & test-B & val-u & test-u \\

\midrule

\multicolumn{9}{c}{\it Specialist models} \\

\midrule

\color{gray} UNINEXT & \color{gray} 85.24 & \color{gray} 89.63 & \color{gray} 79.79 & \color{gray} 92.64 & \color{gray} 94.33 & \color{gray} 91.46 & \color{gray} 88.73 & \color{gray} 89.37 \\
\color{gray} G-DINO-L & \color{gray} 82.75 & \color{gray} 88.95 & \color{gray} 75.92 & \color{gray} 90.56 & \color{gray} 93.19 & \color{gray} 88.24 & \color{gray} 86.13 & \color{gray} 87.02\\

\midrule

\multicolumn{9}{c}{\it Generalist models} \\

\midrule

OFA-L & 68.29 & 76.00 & 61.75 & 79.96 & 83.67 & 76.39 & 67.57 & 67.58 \\
Shikra 13B & 82.89 & 87.79 & 74.41 & 87.83 & 91.11 & 81.81 & 82.64 & 83.16 \\
MiniGPT-v2-7B & 79.97 & 85.12 & 74.45 & 88.69 & 91.65 & 85.33 & 84.44 & 84.66 \\
\begin{tabular}{@{}l@{}}MiniGPT-v2-7B\\\ \ \ \ \ \ \ \ \ \ \ \ \ \ \ \ \ \ -Chat\end{tabular} & 79.58 & 85.52 & 73.32 & 88.06 & 91.29 & 84.30 & 84.19 & 84.31 \\
Qwen-VL-7B & 83.12 & 88.25 & 77.21 & 89.36 & 92.26 & 85.34 & 85.58 & 85.48 \\
\begin{tabular}{@{}l@{}}Qwen-VL-7B\\\ \ \ \ \ \ \ \ \ \ \ \ \ \ \ \ \ \ -Chat\end{tabular} & 82.82 & 88.59 & 76.79 & 88.55 & 92.27 & 84.51 & 85.96 & 86.32 \\

SPHINX & 86.64 & 91.08 & 80.35 & 91.05 & 92.65 & 86.56 & 88.19 & 88.35 \\
\midrule
\rowcolor[gray]{0.95}
\multicolumn{1}{l|}{\cellcolor[gray]{0.95}SPHINX-Tiny} & 71.34	&78.49	&63.71& 82.89&	86.89&	77.91	&	78.50&	78.86 \\
\rowcolor[gray]{0.95}
\multicolumn{1}{l|}{\cellcolor[gray]{0.95}SPHINX-Intern2} & 76.80 &	84.86 &	69.01 &	86.08&	89.70&	81.78&	83.99 &	83.40  \\
\rowcolor[gray]{0.95}
\multicolumn{1}{l|}{\cellcolor[gray]{0.95}SPHINX-Plus} & \textbf{87.59}	& \textbf{92.08}	&\textbf{82.96} &\textbf{92.44}&\textbf{94.22}	&\textbf{90.06}&		\textbf{90.11}&	\textbf{90.56}  \\
\rowcolor[gray]{0.95}
\multicolumn{1}{l|}{\cellcolor[gray]{0.95}SPHINX-MoE} & 85.50	&90.48	&79.88& 90.64&	93.74&	86.85	&	88.26	&88.51 \\

\bottomrule
\end{tabular}
} 
\label{table:ref}
\end{table}


\begin{table*}[t]
    \centering
    \caption{\textbf{Evaluation results of SPHINX-MoE on other MLLM benchmarks.}}
    \vspace{2mm}
    \label{tab:sphinx-moe}
    \adjustbox{max width=\textwidth}{
    \begin{tabular}{l|c|ccccc|cccc|c|cc|ccc}
    \toprule
    \multirow{2}{*}{Methods}&\multirow{2}{*}{MMVP}&\multicolumn{5}{c|}{HallusionBench}&\multicolumn{4}{c|}{AesBench}&\multirow{2}{*}{MMMU-val}&\multicolumn{2}{c|}{CMMMU}&\multicolumn{3}{c}{ScreenSpot}\\
     &&$qAcc$&$fAcc$&Easy $aAcc$&Hard $aAcc$&$aAcc$&AesP&AesE&AesA&AesI&&val&test&Mobile&Desktop&Web \\
     \toprule
     \multicolumn{17}{c}{\textit{Open Source MLLM}} \\
     \midrule
     LLaVA &6.0&-&-&-&-&-&62.43&64.68&45.96&1.125&-&-&-&-&-&-\\
     MiniGPT-4 &12.7&8.79&10.12&31.87&27.67&35.78&41.93&39.35&38.57&0.999&26.8&-&-&6.4&3.7&2.8\\
     InstructBLIP &16.7&9.45&10.11&35.60&\textbf{45.12}&45.26&54.29&53.89&46.54&1.126&32.9&-&- &-&-&-\\
     LLaVA-v1.5 &24.7&10.55&24.86&49.67&29.77&46.94&66.32&68.32&45.46&1.157&36.4&-&- &-&-&-\\
     CogAgent-Chat &-&-&-&-&-&-&-&-&-&-&-&24.6&23.6&46.6&46.5&\textbf{49.7}\\
     Qwen-VL-7B-Chat &-&5.93&6.65&31.43&24.88&39.15&63.21&64.18&46.25&1.192&35.9&30.7&31.3&6.9&5.9&0\\
     \midrule
     \multicolumn{17}{c}{\textit{Closed Source MLLM}} \\
     \midrule
     Gemini-Pro &40.7&-&-&-&-&-&71.99&71.37&49.38&1.222&47.9&-&-&-&-&- \\
     GPT-4V &38.7&\textbf{28.79}&\textbf{39.88}&\textbf{75.60}&37.67&\textbf{65.28}&72.08&70.16&\textbf{50.86}&\textbf{1.301}&\textbf{56.8}&\textbf{42.5}&\textbf{43.7}&-&-&-\\
     \midrule
     \rowcolor[gray]{0.95}SPHINX-MoE&\textbf{49.3}&16.48&23.12&55.16&37.91&52.08&\textbf{72.93}&\textbf{73.32}&49.93&1.267&31.1&29.3&29.6&\textbf{55.1}&\textbf{50.5}&37.3\\
     \bottomrule
    \end{tabular}}
\vspace{-0.3cm}
\end{table*}

\noindent \textbf{MLLM Benchmarks.}
We evaluate SPHINX-X on recently introduced benchmarks, such as MME~\citep{Fu2023MMEAC}, Seedbench~\citep{Li2023SEEDBenchBM}, POPE~\citep{Li2023EvaluatingOH}, LLaVA-Bench (In-the-Wild)~\citep{liu2023llava}, MM-Vet~\citep{Yu2023MMVetEL}, MathVista~\citep{Lu2023MathVistaEM}, MMbench~\citep{liu2023mmbench}, CCbench~\citep{2023opencompass}, Tiny LVLM~\citep{shao2023tiny} and BenchLLM~\cite{Cai2023BenchLMMBC}, InfiMM-Eval~\cite{Han2023InfiMMEvalCO}, Qbench~\cite{cai2023benchlmm} for multi-modal language models (MLLM) to provide a comprehensive assessment of its characteristics. The results, presented in Table~\ref{table:mm}, showcase SPHINX-X's state-of-the-art performance across various multi-modal tasks, including mathematical reasoning, complex scene understanding, low-level vision tasks, and visual quality assessment, as well as resilience when facing illusions.

\noindent \textbf{Visual Question Answering.} The evaluation on general visual question answering (VQA) benchmarks such as VQAV2~\cite{Agrawal2015VQAVQ}, GQV~\cite{Hudson2019GQAAN}, OK-VQA~\cite{Marino2019OKVQAAV}, VizWiz~\cite{Gurari2018VizWizGC}, ScienceQA~\cite{Lu2022LearnTE}, IconQA~\cite{Lu2021IconQAAN} are presented in Table~\ref{table:vqa}. SPHINX-X excels across diverse visual question-answering benchmarks, showcasing its state-of-the-art performance in general visual understanding, relational reasoning, scientific contexts, and symbolic visual reasoning. Moreover, we conduct experiments on text-oriented VQA benchmarks such as TextVQA~\cite{TextVQA}, OCRVQA~\cite{Mishra2019OCRVQAVQ}, DocVQA~\cite{mathew2021docvqa}, ChartQA~\cite{masry-etal-2022-chartqa}, AI2D~\cite{AI2D}, DeepForm~\cite{deepform}, InfoVQA~\cite{Mathew2021InfographicVQA}, TabFact~\cite{Chen2019TabFactAL}, VisualMRC~\cite{Tanaka2021VisualMRCMR}. As shown in Table~\ref{table:textvqa}, SPHINX-X achieves competitive performance on text-related benchmarks with a limited portion of OCR data.

\paragraph{Visual grounding.}
To evaluate SPHINX-X's ability to precisely locate and comprehend referred objects or regions within images, we conduct experiments on Referring Expression Comprehension (REC) benchmarks, including RefCOCO~\citep{Kazemzadeh2014ReferItGameRT}, RefCOCO+~\citep{Mao2015GenerationAC}, and RefCOCOg~\citep{Mao2015GenerationAC}. The results are presented in Table~\ref{table:ref}, SPHINX-X consistently outperforms the majority of state-of-the-art models, surpassing even specialist model G-DINO-L~\cite{Liu2023GroundingDM} and other visual-language generalist models.

\begin{table*}[t]
\centering
\caption{\textbf{Comparison with state-of-the-art methods on Video-Bench.} `\textsuperscript{*}' denotes the QA-pairs are re-constructed or annotated by Video-Bench. `\textsuperscript{V}' denotes a video training version of the model used.}
\vspace{2mm}
\label{table:video_bench}
\adjustbox{max width=\textwidth}{%
\setlength{\tabcolsep}{2.pt}
\begin{tabular}{@{}l|c|ccccccc|ccc|ccc@{}}
\toprule
\multirow{2}{*}{{Methods}} & \multirow{2}{*}{{Avg.}} &\multicolumn{7}{c}{{Video-Exclusive Understanding}} &\multicolumn{3}{|c|}{{Prior Knowledge-based QA}} &\multicolumn{3}{c}{{Comprehension and Decision-Making}} \\ 
\cmidrule(lr){3-9} \cmidrule(lr){10-12} \cmidrule(lr){13-15}
& & Activitynet-QA & MSVD-QA\textsuperscript{*} & MSRVTT-QA\textsuperscript{*} & TGIF-QA & YouCook2\textsuperscript{*} & UCF-Cirme\textsuperscript{*} & MOT\textsuperscript{*} & TV-QA\textsuperscript{*} & MV-QA\textsuperscript{*}  & NBA-QA\textsuperscript{*} & License Exam\textsuperscript{*} & Decision-Making\textsuperscript{*} & SQA3D\textsuperscript{*} \\ 
\midrule
\multicolumn{15}{c}{\it Video-based MLLM} \\ \midrule
Video-LLaMA & 31.8 & 39.9 & 41.2 & 34.1 & 31.3 & 28.9 & 27.6 & 16.7 & 24.8 & 32.4 & 26.2 & 30.6 & 49.1 & 31.2 \\
mPLUG-Owl\textsuperscript{V}  & 32.7 & 41.5 & 42.5  & 36.3  & 31.7  & 27.1 &  22.8  & \textbf{27.8}  & 24.0  & 30.2 & 25.1 & 33.3 & 51.0 & 32.0  \\
VideoChat & 34.6 & 44.6 & 42.2 & 37.4 & 33.7 & 27.7  & 22.4  & \textbf{27.8} & 26.2 & 34.1 & 28.6 & 38.9 & 55.4  & 31.4 \\
Chat-UniVi & 35.2 & 49.0 & 48.6 & 41.7  & 41.3 & 29.0  & \textbf{28.3}  & 16.7 & 23.1 & 33.6 & 25.7 & 38.9 & 53.1 & 29.1 \\ 
PandaGPT & 36.7 & 45.0 & 50.4 & 44.6 & 29.7 & 33.0  & 33.0  & 16.7 & 27.9 & 37.1 & 31.1 & 41.7 & 56.0 & 30.8 \\
Otter\textsuperscript{V}  & 37.1 & 44.3 & 55.0 & 47.0 & 34.3 & 32.7  & 22.4  & 16.7 & 27.7 & 37.1 & 34.3 & \textbf{52.8} & 48.7 & 29.7 \\
Video-ChatGPT & 38.3 & 46.6 & 57.5 & 46.3  & 35.6 & 34.8  & 24.1 & \textbf{27.8} & 28.8 & 36.5 & 22.5 & 41.7 & \textbf{58.2} & 37.2 \\ \midrule
\multicolumn{15}{c}{\it Image-based MLLM} \\ \midrule
SPHINX & 39.0 & 50.1 & 56.7 & 45.4 & 42.8 & 37.0 & 25.2 & 5.6 & 29.8 & 33.3 & 30.9 & 50.0 & 52.8 & 47.7 \\
\midrule
\rowcolor[gray]{0.95}
SPHINX-Plus & \textbf{45.1} & \textbf{53.1} & \textbf{68.5} & \textbf{54.0} & \textbf{53.4} & \textbf{42.0} & 27.6 & 11.1 & \textbf{36.5} & \textbf{44.0} & \textbf{45.0} & 47.2 & 55.6 & \textbf{48.8} \\
\bottomrule
\end{tabular}
} 
\end{table*}


\subsection{SPHINX-MoE on other MLLM Benchmarks}
To investigate the ability of SPHINX-MoE more concretely and locate its ability level among many developed MLLMs, we evaluate SPHINX-MoE on some recently curated benchmarks, which are listed below:

\vspace{-1em}
\begin{itemize}
\small
\setlength{\itemsep}{-1pt}
\item[-] \textbf{MathVerse}~\cite{zhang2024mathverse}: A mathematical benchmark in visual contexts to explore the multi-modal diagram interpretation and reasoning capabilities of MLLMs, which annotate math problems into different versions for fine-grained evaluation.
\item[-] \textbf{SciVerse}~\cite{sciverse}: A comprehensive scientific problem benchmark (physics, chemistry, and biology) in visual contexts to reveal the domain-specific knowledge comprehension proficiency of MLLMs.
    \item[-] \textbf{MMVP}~\cite{tong24mmvp}: A benchmark specially crafted to measure MLLMs' visual understanding capability.
    \item[-] \textbf{HallusionBench}~\cite{guan2023hallusionbench}: A benchmark to agnostic MLLMs' language hallucination and visual illusion.
    \item[-] \textbf{AesBench}~\cite{AesBench}: An expert benchmark aiming to comprehensively evaluate the aesthetic perception capacities of MLLMs.
    \item[-] \textbf{MMMU}~\cite{yue2023mmmu} \& \textbf{CMMMU}~\cite{zhang2024cmmmu}: An English and a Chinese benchmark, respectively, aiming to solve massive multi-discipline tasks, which need college-level subject knowledge and deliberate reasoning ability.
    \item[-] \textbf{ScreenSpot}~\cite{cheng2024seeclick}: A benchmark across various GUI platforms and designed to assess MLLM's capability to localize elements based on the human's instructions.
\end{itemize}
\vspace{-1em}

The results for MathVerse and SciVerse are showcased in Table~\ref{tab:ms}. Our SPHINX-MoE attains the best performance among open-source models, indicating superior math problem-solving and scientific understanding capabilities.
The results of our model on other benchmarks are included in Table~\ref{tab:sphinx-moe}. As we can see SPHINX-MoE performs well on all benchmarks, so we can infer that (i) SPHINX-MoE has a better visual understanding ability and less language hallucination than other competitors. (ii) SPHINX-MoE can deal with the web and mobile domain data well. It should be noted that on some tasks or metrics, SPHINX-MoE performs even better than GPT-4V, \emph{e.g.}, MMVP and AesP, AesE in AesBench. However, it is hard for SPHINX-MoE to solve the multi-discipline tasks, \emph{i.e.}, the MMMU and CMMMU benchmark. And we think this is due to the lack of multi-modal multi-disciplinary data during the training stage. Thus we would consider to involving more multi-disciplinary data in SPHINX-MoE's training.

\subsection{Performance of SPHINX-Plus on Video Analysis}
To further assess the visual comprehension capabilities of our method, we conduct additional experiments on challenging video tasks. Since SPHINX-Plus is an image-based MLLM and is not trained on any video data, we need to do additional processing on video inputs. To be specific, we evenly sampled videos and selected the middle frame as the representative frame for input into the model. We conduct extensive experiments on 
Video-Bench~\citep{ning2023video} which evaluates the performance of models across three distinct capability levels: (i) video-exclusive understanding, (ii) prior knowledge-based question-answering, and (iii) comprehension and decision-making.

As shown in \cref{table:video_bench}, SPHINX-Plus, despite being an image-based model, significantly outperforms existing models~\citep{jin2023chat,su2023pandagpt,zhang2023video} specifically tailored for video tasks. Especially in the aspects of video-exclusive understanding and prior knowledge-based question-answering, SPHINX-Plus showcases outstanding performance, signifying its proficiency in visual perception and knowledge extraction capabilities. In challenging datasets such as MOT, SPHINX-Plus demonstrates slightly lower performance compared to existing state-of-the-art methods~\citep{maaz2023video,li2023videochat}. We attribute this to the need to model timing relationships in videos. SPHINX-Plus has not been fine-tuned by any video data, so its performance marginally underperforms others.

\subsection{Demonstrations of SPHINX-X}

In Figure~\ref{fig:fig2}, the demonstrates of SPHINX-X indicate that our models can 1) conduct fine-grained object detection in high-resolution images by the proposed sub-image division strategy; 2) conduct text spotting with accurate content and positions; 3) engage in bilingual image-based conversations, and generate coherent, accurate, and detailed Chinese descriptions for synthetic images; 4) generate accurate code for visual programming based on the precise understanding of the given plot; 5) analyze App screenshots based on the functional description and output the corresponding bounding box; 6) accurately interpret geometric questions from images, thanks to the math and extensive OCR datasets included in our training corpus; and 7) estimate the correct pose with rigorous body key points.

In the appendix (Figure~\ref{fig:fig3}), we respectively show the Set-of-marks (SoM) prompting and OCR understanding capabilities of SPHINX-X. With our curated SoM dataset, SPHINX-X can well understand the marks given in the prompt, i.e., a cat and bear bottle, and analyze the appearance and relations of designated objects. By the training of OCR-intensive data, our model can conduct accurate document layout detection and character recognition.

\section{Conclusion}

In this paper, we introduce SPHINX-X, a series of MLLMs for multi-purpose multi-modal instruction tuning with LLM parameters ranging from 1B to 8$\times$7B. On top of the original SPHINX, we propose three aspects of improvements, i.e., removing redundant visual encoders, bypassing fully-padded sub-images with skip tokens, and simplifying multi-stage training into a one-stage all-in-one paradigm. We also curate a large-scale multi-domain dataset for MLLM training, which contains a wide range of public datasets and our constructed targeted data. Extensive benchmarks and evaluations demonstrate the superior performance and generalization capacity of SPHINX-X. We hope our work may cast a light on future MLLM research.
\clearpage








\section*{Impact Statement}

The SPHINX-X Multi-modality Large Language Model series has the potential to impact society in several ways:

\textbf{Enhanced Multimodal AI Applications}: SPHINX-X could lead to the development of more sophisticated AI systems capable of understanding and interacting with both text and visual input. This can improve services like automated translations, image recognition, and assistive technologies for individuals with disabilities.

\textbf{Inclusivity and Language Diversity}: By training on a diverse, multi-domain, and multi-modal dataset, the model may offer broader language support, which can bridge communication gaps and foster inclusivity.

\textbf{Ethical and Bias Considerations}: The expansive dataset used for training must be carefully curated to avoid perpetuating biases, ensuring that the model’s responses are fair and ethical.

\textbf{Misuse Risks}: The misuse of MLLMs for generating deceptive content is a risk, underscoring the need for robust governance and ethical usage frameworks.

The responsible deployment of SPHINX-X requires careful consideration of these potential impacts to maximize benefits and minimize negative consequences.





\bibliography{sphinx-x}
\bibliographystyle{icml2024}

\clearpage
\appendix

\begin{figure*}[t]
\vspace{0.3cm}
    \centering
\includegraphics[width=0.9\textwidth]{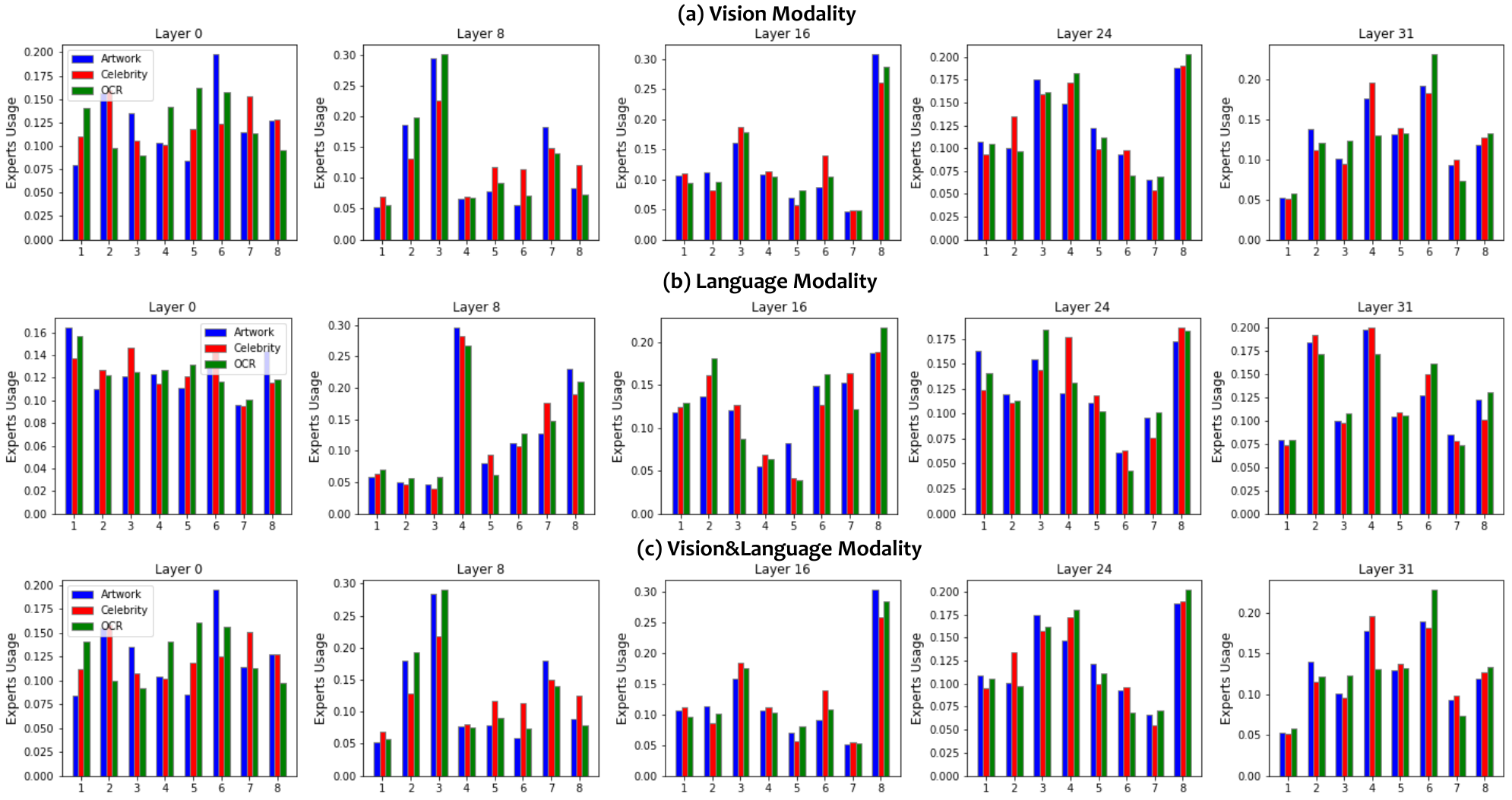}
    \caption{\textbf{Experts' usage distribution on different domains and different modalities.}}
    \label{fig:experts_usage}
\end{figure*}

\section{Appendix}

\begin{table*}[t]
\vspace{-3mm}
\centering
\caption{\textbf{Comparison with state-of-the-art methods on MVBench.} `\textsuperscript{V}' denotes a video training version of the model used.}
\vspace{2mm}
\label{table:MVBench}
\adjustbox{max width=\textwidth}{%
\setlength{\tabcolsep}{3pt}
\begin{tabular}{@{}l|c|cccccccccccccccccccc@{}}
\toprule
{{Methods}} & {{Avg.}} & AS & AP & AA & FA & UA & OE & OI & OS & MD & AL & ST & AC & MC & MA & SC & FP & CO & EN & ER & CI \\ 
\midrule
\multicolumn{22}{c}{\it Video-based MLLM} \\ \midrule
Otter\textsuperscript{V}  & 26.8 & 23.0 & 23.0 & 27.5 & 27.0 & 29.5  & 53.0 & 28.0 & 33.0 & 24.5 & 23.5 & 27.5 & 26.0 & 28.5 & 38.5 & 22.0 & 18.0 & 22.0 & 23.5 & 19.0 & 19.5 \\
mPLUG-Owl\textsuperscript{V}  & 29.7 & 22.0 & 28.0  & 34.0 & 29.0 & 29.0 &  40.5 & 27.0 & 31.5 & 27.0 & 23.0 & 29.0 & 31.5 & 27.0 & 44.0 & 24.0 & 40.0 & 31.0 & 26.0 & 20.5 & 29.5 \\
Video-ChatGPT & 32.7 & 23.5 & 26.0 & 62.0 & 22.5 & 26.5 & 54.0 & 28.0 & 40.0 & 23.0 & 20.0 & 31.0 & 30.5 & 25.5 & 48.5 & 29.0 & 39.5 & 33.0 & 29.5 & 26.0 & 35.5 \\
Video-LLaMA & 34.1 & 27.5 & 25.5 & 51.0 & 29.0 & 39.0 & 48.0 & 40.5 &  38.0 & 22.5 & 22.5 & 43.0 & 34.0 &  22.5 & 45.5 & 32.5 & 32.5 & 40.0 & 30.0 & 21.0 & 37.0 \\
VideoChat & 35.5 & 33.5 & 26.5 & 56.0 & 33.5 & 40.5 & 53.0 & 40.5 & 30.0 & 25.5 & \textbf{27.0} & 48.5 & 35.0 & 20.5 & 46.0 & 26.5 & 42.5 & \textbf{41.0} & 23.5 & 23.5 & 36.0 \\
VideoChat2 & \textbf{51.1} & \textbf{66.0} & \textbf{47.5} & \textbf{83.5} & \textbf{49.5} & \textbf{60.0} & \textbf{58.0} & \textbf{71.5} & \textbf{42.5} & 23.0 & 23.0 & \textbf{88.5} & \textbf{39.0} & \textbf{42.0} & 44.0 & \textbf{49.0} & \textbf{58.5} & 36.5 & \textbf{35.0} & \textbf{40.5} & \textbf{65.5} \\ \midrule
\multicolumn{22}{c}{\it Image-based MLLM} \\ \midrule
SPHINX & 37.5 & 32.5 & 31.5 & 65.0 & 38.5 & 43.5 & 54.0 & 37.5 & 28.5 & 22.5 & 26.5 & 45.5 & \textbf{39.0} & 41.0 & 47.5 & 40.0 & 23.5 & {37.5} & {31.0} & {35.0} & 30.5 \\
\midrule
\rowcolor[gray]{0.95}
SPHINX-Plus & {39.7} & {47.5} & {32.0} & {58.0} & {42.5} & {43.5} & 45.0 & 44.0 & {35.5} & \textbf{29.0} & \textbf{27.0} & 52.0 & 38.0 & {41.0} & \textbf{59.5} & 37.5 & 23.0 & \textbf{41.0} & 29.0 & 40.0 & 29.5 \\
\bottomrule
\end{tabular}
} 
\vspace{0.2cm}
\end{table*}

\subsection{Analysis of Routing Mechanisms in SPHINX-MoE}
\subsubsection{Inference with different numbers of activating experts}

\begin{figure}[h]
    \vspace{0.2cm}
    \centering
    \includegraphics[width=\linewidth]{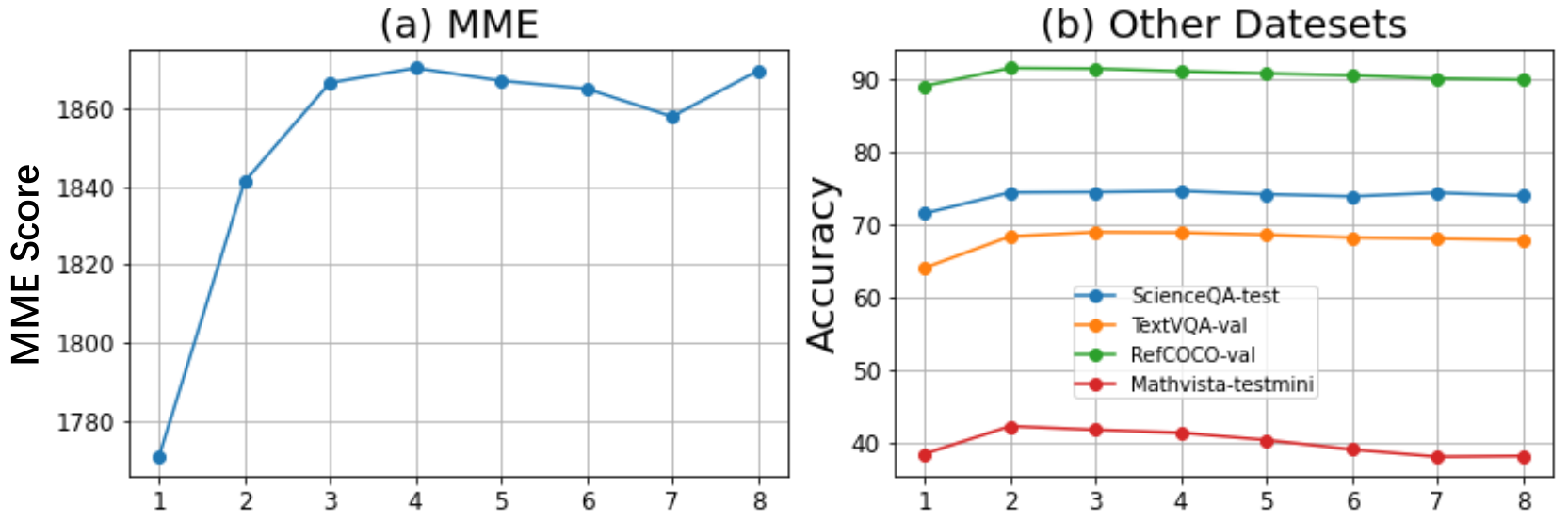}
    \vspace{-0.4cm}
    \caption{\textbf{Performance with different numbers of activating experts when inference.} We respectively report the performance on MME and other benchmarks.}
    \vspace{-0.1cm}
    \label{fig:aen}
\end{figure}

For SPHINX-MoE, the LLM backbone is based on Mixtral-8$\times$7B~\cite{albert24mixtral}, which is a mixture-of-experts-based large language model. Thus, during the inference time, only some of the experts will be activated when dealing with each token. In the training stage of SPHINX-MoE, only two of the eight experts will be activated, so we set the default number of activating experts to 2 when inference. To investigate how the activating experts' amount will affect the inference performance, we change it from one to eight, and the results are shown in Figure~\ref{fig:aen}. 

As we can see, on most datasets, \emph{i.e.}, ScienceQA~\cite{Lu2022LearnTE}, TextQA~\cite{TextVQA}, RefCOCO~\cite{Lin2014MicrosoftCC} and Mathvista~\cite{Lu2023MathVistaEM}, when activating two experts, keeping it the same with training setting, SPHINX-MoE performs the best. However, for MME~\cite{fu2023mme}, when setting the number of activating experts to four, SPHINX-MoE works the best. Two activating experts actually make the second low-performance. This inconsistency with the training setting is interesting.

\subsubsection{Experts' usage distribution on different domains and different modalities}

In some previous works, each expert in the mixture-of-experts model is a specialist for a specific domain or modality, \emph{e.g.}, VLMO~\cite{bao22vlmo}. So we explore that, in SPHINX-MoE, how each expert in each layer deals with data from different domains and different modalities. So we pick the artwork, celebrity and OCR subtasks from the MME~\cite{fu2023mme} benchmark, and infer SPHINX-MoE on these subtasks with two activating experts, recording the expert's usage distribution of each layer, as shown in Figure~\ref{fig:experts_usage}. Subfigure (a), (b) and (c) show the results on vision modality, language modality and vision\&language modalities separately. From the distribution record, we don't see an obvious pattern that experts are specialists for different domains or modalities. (i) For different domains, the experts' usage is similar for the three different domain data: artwork, celebrity and OCR. (ii) For different modalities, there are no specific experts that mainly deal with one specific modality. But there is an interesting scenario that the experts' usage distribution of the layers at both ends of the model is more flat than that of the the middle layers.

\subsubsection{Prune some of the experts when inference}

\begin{figure}[h]
    \centering
    \includegraphics[width=0.9\linewidth]{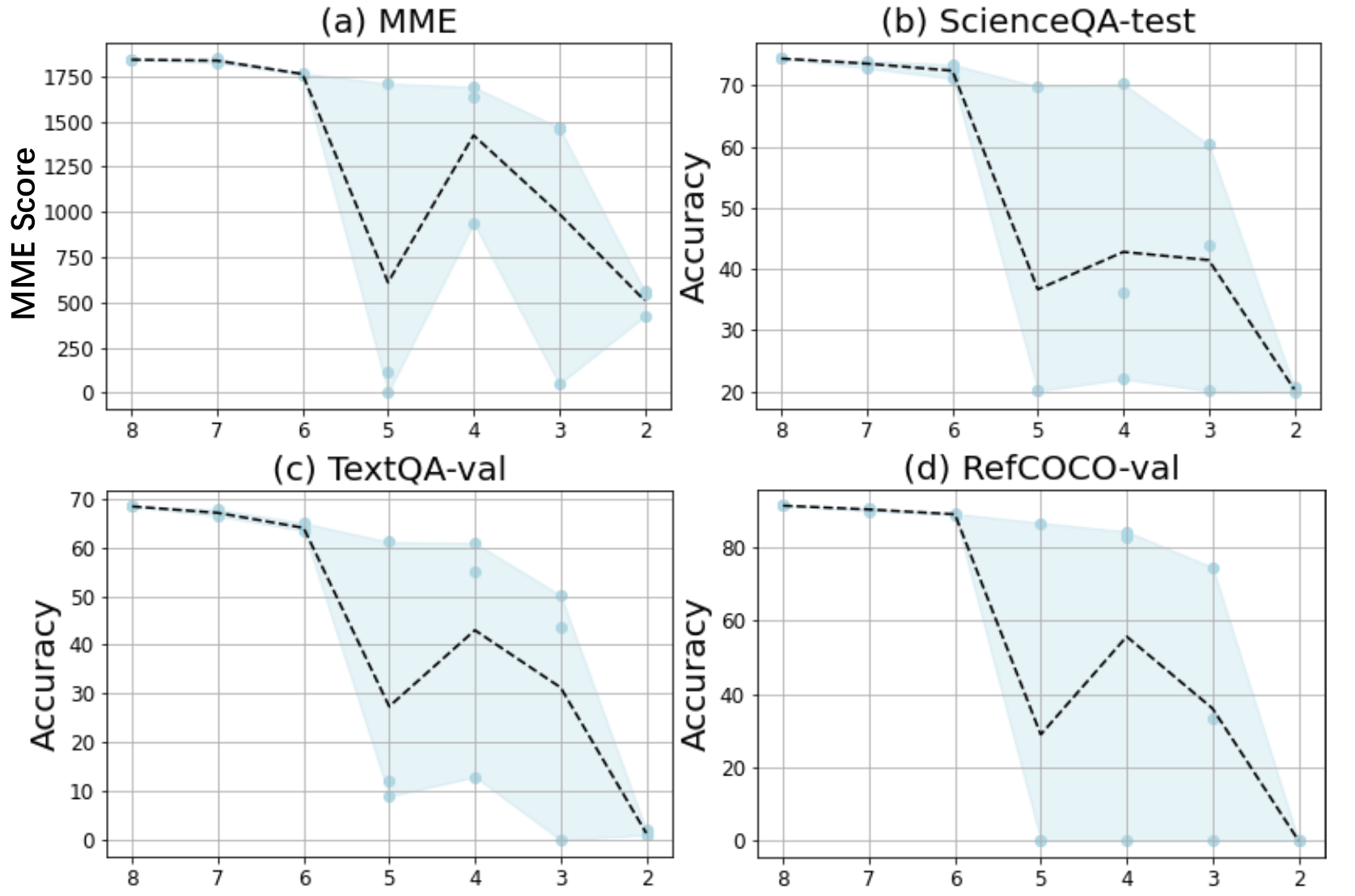}
    \caption{\textbf{Performance change when pruning different numbers of experts in each layer.} The black dotted line is the average of the three runs of random pruning.}
    \label{fig:prune}
\end{figure}

Different from the dense model, for the sparse model only part of the parameters are activated during the inference time. So if we prune some experts, the ability of the model could be partly saved. To investigate how the number of pruned experts will affect SPHINX-MoE's ability, we prune different numbers of experts of SPHINX-MoE's each layer, and the results are shown in Figure~\ref{fig:prune}. In Figure~\ref{fig:prune}, the x-axis means the number of retained experts after pruning. For each value on the x-axis, termed as ``$n$", we randomly choose $8-n$ experts in each layer to be pruned in SPHINX-MoE. For each $n$, we run it three times for average performance. 

We find that some experts are ``important'', \emph{i.e.}, there is a huge performance variance in the three runs. If we prune some specific experts, SPHINX-MoE will lose most ability, while in another run, the most ability of the model will be retained even if we prune the same number of experts in each layer. Thus, if we keep these ``important'' experts and prune other less ``important'' experts in SPHINX-MoE, most ability could be saved, as the upper tendency in Figure~\ref{fig:prune}.

\subsection{Video Analysis on MVBench}

To further evaluate the video understanding capacity, we evaluate SPHINX-X on MVBench~\citep{li2023mvbench}, which breaks down video understanding into 20 sub-aspects, allowing for a more detailed comparison of model performance at a finer granularity.
As shown in \cref{table:video_bench}, SPHINX-Plus, despite being an image-based model, significantly outperforms existing models~\citep{jin2023chat,su2023pandagpt,zhang2023video} specifically tailored for video tasks. Especially in the aspects of video-exclusive understanding and prior knowledge-based question-answering, SPHINX-Plus showcases outstanding performance, signifying its proficiency in visual perception and knowledge extraction capabilities. In challenging datasets such as MOT, SPHINX-Plus demonstrates slightly lower performance compared to existing state-of-the-art methods~\citep{maaz2023video,li2023videochat}. We attribute this to the need to model timing relationships in videos. SPHINX-Plus has not been fine-tuned by any video data, so its performance marginally underperforms others.

\subsection{Additional details on the training dataset}
\label{sec:train_data_appendix}

\noindent \textbf{Language Instruction-following Data.} 
Unlike previous works~\cite{zhu2023minigpt,liu2023llava,liu2023improvedllava} that utilize instruction-tuned LLMs such as Vicuna~\cite{vicuna2023}, SPHINX-X is directly trained on top of the basic pre-trained LLM, i.e., LLaMA2~\cite{touvron2023llama2}. This is to investigate the training characteristics of multi-modal models from LLMs more clearly. Therefore, we are required to collect a high-quality dataset combination for language instruction-following. The dataset includes multi-turn dialog, question-answering, code generation, and math word problems. In detail, UltraChat~\cite{ding2023enhancing} and OpenOrca~\cite{OpenOrca} are utilized for basic multi-turn conversation abilities. MetaMath~\cite{yu2023metamath} and MathInstruct~\cite{yue2023mammoth} are high-quality mathematical datasets with reasoning process. WizardCoder~\cite{luo2023wizardcoder} is adopted for increasing the coding ability of LLMs. Flan-mini~\cite{ghosal2023flacuna} is a subset of FLAN datasets and is included for question-answering capabilities.

\noindent \textbf{Visual Instruction-following Data.}
For comprehensive visual understanding, we expand the data scale of SPHINX to incorporate a variety of vision tasks and transform their annotations into a unified question-answering format.
The tasks include image classification~\cite{Russakovsky2014ImageNetLS}, object detection such as COCO~\cite{Lin2014MicrosoftCC},OpenImages~\cite{Kuznetsova2018TheOI},Object365~\cite{Shao2019Objects365AL},Lvis~\cite{Gupta2019LVISAD}, human pose estimation such as UniPose~\cite{Yang2023UniPoseDA}, COCO-Pose~\cite{Lin2014MicrosoftCC}, and visual grounding. 
We utilize a task-specific prompt as the question, and regard the ground-truth labels as the answer by textualizing them in language space.
For generality, we do not utilize any special tokens for different tasks, and treat them all as pure language problems.
This visual supervised fine-tuning enhances SPHINX-X with the performance of image parsing, object localization, and relation reasoning, empowering MLLMs with in-built capacity to be a universal visual generalist.

\noindent \textbf{Vision-language Instruction-following Data.}
To align MoV with LLMs and enable visual instruction following, we gather large-scale datasets from established visual question-answering sources such as VQAV2~\cite{Agrawal2015VQAVQ}, GQA~\cite{Hudson2019GQAAN}, OK-VQA~\cite{Marino2019OKVQAAV}, Visual Genome~\cite{krishna2017visual}, and CLEVR~\cite{Johnson2016CLEVRAD}. To specifically boost SPHINX-X's text-oriented VQA capabilities, we incorporate datasets including TextVQA~\cite{TextVQA}, DocVQA~\cite{mathew2021docvqa}, ChartQA~\cite{masry-etal-2022-chartqa}, AI2D~\cite{AI2D}, Deepform~\cite{deepform}, DVQA~\cite{Kafle2018DVQAUD}, InfographicsVQ~\cite{Mathew2021InfographicVQA}, KleisterCharity~\cite{Stanislawek2021KleisterKI}, TabFact~\cite{Chen2019TabFactAL}, VisualMRC~\cite{Tanaka2021VisualMRCMR}, and WikiTableQuestions~\cite{Pasupat2015CompositionalSP}. Leveraging the rich knowledge embedded in large foundation models, we also encompass high-quality MLLM-generated data, e.g., dense captioning data of ShareGPT4V~\cite{Chen2023ShareGPT4VIL} and visual instruction data from LLaVA~\cite{liu2023llava}, LVIS-INSTRUCT4V~\cite{lavisinstruct}, and LLaVAR~\cite{Zhang2023LLaVAREV}. Additionally, we employ Geometry3K~\cite{Lu2021InterGPSIG} to enhance the model's geometry problem-solving abilities.

\begin{table*}[t]
\vspace{-0.2cm}
\centering
\scriptsize
\caption{\textbf{One-stage training data summary of SPHINX-X.}}
\vspace{2mm}
\label{tab:data_summary}
\begin{tabular}{l|c|l} 
\toprule
Tasks              & \#Samples & \makecell*[c]{\ \ \ \ \ \ \ \ Datasets} \\ 
\midrule
\rowcolor{gray!10} \multicolumn{3}{l}{Language Instruction-following Data}\\ 
\midrule
Multi-turn Dialog  & 1.8M       & UltraChat,Flan-mini,OpenOrca   \\
Math               & 0.6M       & MetaMathQA,MathInstruct  \\
Coding             & 80k        & WizardCoder \\ 
\midrule
\rowcolor{gray!10} \multicolumn{3}{l}{Visual Instruction-following Data} \\ 
\midrule
Detection          & 4.9M       & V3Det,OpenImages,Lvis,COCO,Object365  \\
Human Pose         & 0.3M       & Unipose,COCO-Pose\\
Classification     & 1M         & ImageNet1K     \\
Grounding          & 1M         & Visual Genome, RefCOCO, RefCOCO+,RefCOCOg,Flickr30k  \\ 
\midrule
\rowcolor{gray!10} \multicolumn{3}{l}{Vision-language Instruction-following Data}\\
\midrule
VQA                & 0.7M       & \begin{tabular}{@{}l@{}}VQAV2,OKVQA,GQA,Visual Genome\\CLEVR,ChartQA,DeepForm,DocVQA\\DVQA, InfographicsVQA,KleisterCharity\\VisualMRC,WikiTableQuestions\\TextVQA,TabFact \end{tabular}  \\
\cmidrule{3-3}
Caption            & 0.5M       & MSCOCO,ShareGPT4V,LaionGPV4V    \\
Visual Instruction & 0.4M       & LLaVA,LVIS-INSTRUCT4V,LLaVAR\\
\midrule                                        
\rowcolor{gray!10} \multicolumn{3}{l}{OCR-intensive Data}   \\ 
\midrule 
OCR    & 3M       & PaperText: Arxiv, Common Crawl\\
Text Layout \& Spotting         &   1.0 M   & DocBank, M6Doc, Publaynet, DocLayNet, ICDAR, CTW1500  \\
\midrule
\rowcolor{gray!10} \multicolumn{3}{l}{Set-of-Marks Instruction-following Data}\\ 
\midrule 
Natural Images                  & 5k       & COCO, LVIS, Visual Genome     \\
Website/Mobile/Desktop agent    & 1k       & SeeClick       \\       
OCR-related     & 2k       &  TotalText,CTW1500,IC13,IC15    \\ 
Document Images     & 1k       & M6Doc, DoclayNet, PublayNet     \\ 
Multipanel Images               & 1k       & In-house dataset
\\ \bottomrule
\end{tabular}
\vspace{-0.1cm}
\end{table*}

\begin{figure*}[t]
    \centering
    \includegraphics[width=\linewidth]{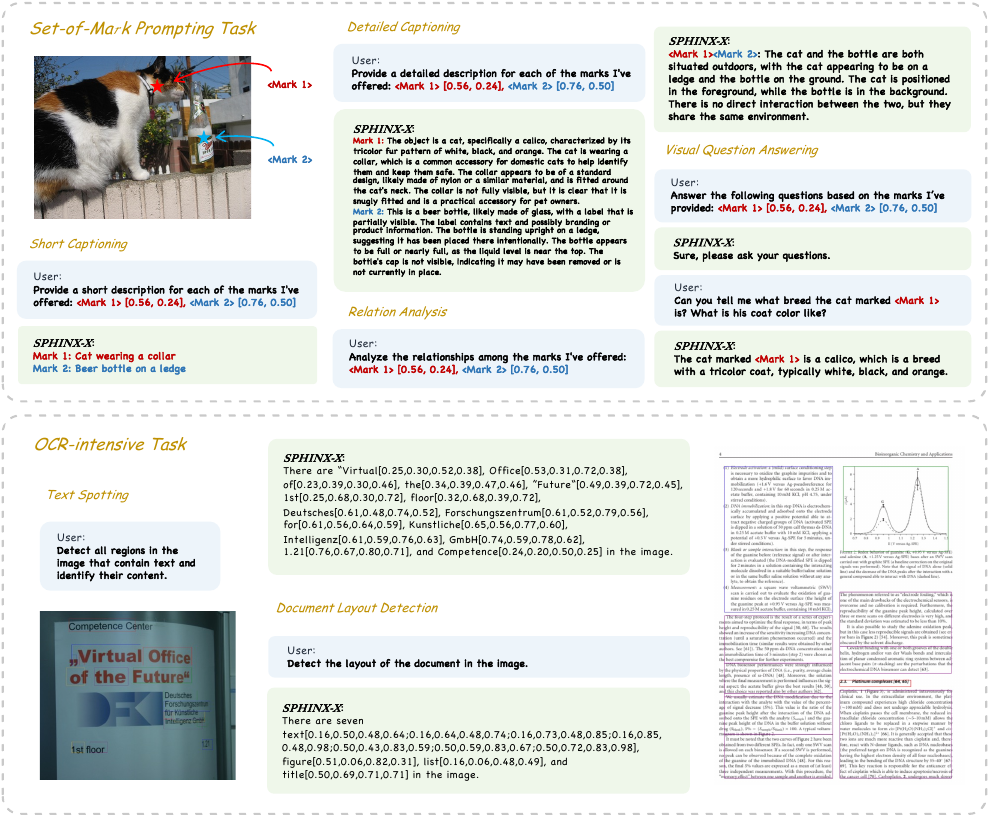}
    \vspace{-0.3cm}
    \caption{\textbf{Set-of-Marks (SoM) prompting and OCR-intensive capabilities of SPHINX-X.} With our constructed two datasets, SPHINX-X exhibits outstanding visual performance on SoM prompting and OCR-related tasks. Note that the SoM marks are only utilized in the textual prompt, without rendering on input images.}
    \label{fig:fig3}
\end{figure*}


\end{document}